\documentclass[a4paper,twoside,openright,twocolumn]{article}

\usepackage[latin1]{inputenc}
\usepackage[pdftex]{graphicx}
\usepackage[numbers]{natbib}
\usepackage{listings}
\usepackage{xcolor}
\usepackage{float}
\usepackage{wrapfig}
\usepackage{amsfonts}
\usepackage[normal]{subfigure}
\usepackage{amsmath}
\usepackage{amsthm}
\usepackage{verbatim}

\title{An overview on automatic design of robot controllers for complex tasks}
\author{Michele Matteini}

\newcommand{\image}[3]{
	\begin{figure}[H]
		\begin{center}
			\includegraphics[scale=#3]{#1.png} 
		\end{center}
		\caption{\textit{#2}\label{fig:#1}}
	\end{figure}
}

\newcommand{\refig}[1]{(see \figurename~\ref{fig:#1})}

\begin{document}
\maketitle
\pagestyle{plain}

\tableofcontents

\section{Introduction}
In this paper we will explore different available methodologies to automatically design controllers for tasks that spans many level of abstraction, where the gap between primitive behaviours and the task definition is high. A good understanding of your evolutionary setup is needed to choose the correct strategy with which to tackle complex tasks thus we'll first review the most used types of each element composing an evolutionary setup (controllers, objective functions, ect.) then we'll move the focus on the bootstrapping problem and on the different strategies used to overcome it.

\section{Automatic design solutions}

\image{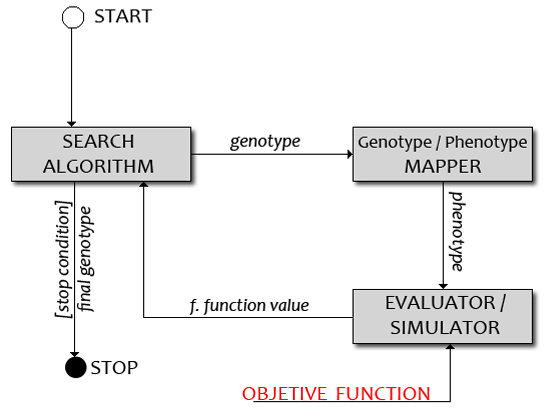}{Interactions between elements of an evolutionary setup: arches rappresent input/output. In red: the objective function as an input for the evaluation stage, together with the phenotype (controller)}{0.4}

When it comes to automatic design, there are many elements to take into consideration: a configurable or evolvable \textit{controller} is needed, on which an \textit{algorithm} that find the best configuration for that controller is applied. The algorithm must be able to evaluate how good each configuration perform on the required task, for this purpose an \textit{objective function} is used. Sometimes a more convenient and/or compact rappresentation of the controller's evolvable parameters can be used in the evolutive process (\textit{genotype}). A \textit{mapping} can then be defined to obtain the controller corrisponding to that rappresentation (\textit{phenotype}). 
\\
With all this elements in place we'll have defined an evolutionary setup \refig{ev-setup}. For each one of these (controller, algorithm, fitness function\footnote{``finess function'' usually refers to a type of objective function used in GAs but since here this differentiation is not needed, this term will always refer to an objective function.}, genotype-phenotype mapping) there are however many possible implementations that will be explored in the sections below. 

\subsection{Evolvable Controllers}
There are many controllers that can be evolved with an algorithm, some of them are usable only in the ER field, others can be also designed manually. Among the most used, we can find:

\begin{itemize}
	\item \textbf{Artificial Neural Networks (ANN)}\\
	These are the most used evolvable controllers due to their fast convergence and the complexity of the tasks that can be learned. Algorithms have been developed that can train this kind of network in a very short time (\citet{art003}) that aren't task-specific and can be used on many tasks (\citet{art010}). The variable parameters in the evolution process are the weights of the arches, the threasholds of neurons and in some cases even the network topology. The downside of choosing ANNs is mainly the fact that once you obtain a solution, you dont know how it works internally thus making impossible to test properties or doing any kind of analisys (except on very small networks) and can only be used as black-boxes.
	
	\item \textbf{Programs (executable code)}\\
	Programs composed by executable instructions can also be used as evolvable controllers: once an instruction set has been defined, many algoritms can be used to search on the space of all the possible programs of a given lenght. As an example in \citet{art020} work a program evolved with a genetic algorithm is used to generate a walking loop on different robot morphologies.

	\item \textbf{Finite State Machines (FSM)}\\
	Attempts on the evolution of FSMs have been done in \citet{art001} with the main benefit of the solution being human-readable. Using an FSM means that you can easily combine primitive parametric behaviours by considering them as states in the automations.
	
	\item \textbf{Random Boolean Networks (RBN)}
	This kind of network have been used only recently as controllers in ER. Evolving this networks can produce complex behaviours comparable to the ones obtained with ANN but with a limited state space size, that leave open the possibilities for analysing their internal behaviours. Some studies have been done(see \citet{art012} and \citet{art013}) that suggest that RBNs behaviour can ba analized by mapping them to probabilistic FSM.
	The disadvantage of using RBNs lies in the limitations for I/O: input from the sensors needs an encoding step before being injected as boolean values in the network, and the same is true for outputs that are expressed as booleans and requires a mapping to the actuators; this can be a problem if the task require complex I/O values. 
	
\end{itemize}

\subsection{Fitness functions}
A core element of the evolutionary cycle illustrated in figure \ref{fig:ev-setup} is the fitness function which once evaluated, discriminates the \textit{``good''} controllers from the low performing ones. The training of a robot for a given task entirely depends on the formulation of the right fitness function that can select the most successful controllers without including task-specific aspects that may introduce a bias in the process.
A suggested classification for fitness functions in ER is based on the degree of \textit{``a priori knowledge''} (see \citet{art007}) and from the most to the least task-dependent, can be summarized as follows:
\begin{itemize}
	
	\item \textbf{Training data fitness functions}\\
	These are the kind of fitness functions that measure errors on datasets containing example of task instances coupled with the expected output. Using these functions mean that the task is almost completely defined and the robot is not going to discover anything new, so they're better suited for classification tasks or where the robot have to mimic the behaviour of a human. 
	
	\item \textbf{Behavioral fitness functions}\\
	Task-specific functions that measures how good the robot is doing by evaluating key sensing/acting reflexes useful for the task (e.g. in obstacle avoidance, a term that give an higher score to robots that turns when front proximity sensors are stimulated). These are often composed of many terms (one for each needed behaviour), combined in a weighted sum.
	
	\item \textbf{Functional incremental fitness functions}\\
	Used in incremental evolution where simpler or primitive behaviors are learned first, and the evolutive path is assisted by the fitness function that change through the evolution until reaching the one that evaluates the complete task. These usually requires an \textit{a priori knowledge} similar or lower than behavioral ones, but the course of evolution is restricted resulting in more trivial solutions and less novelty.
	
	\item \textbf{Tailored fitness functions}\\
	The function measure the degree of completion of the task, thus requiring less knowledge about possible solutions, letting the robots free to explore all the possibilities
	(e.g. for a phototaxis behaviour, the distance separating the robot from the light should be minimized like in \citet{art009}).

	\item \textbf{Environmental incremental fitness functions}\\
	Similar to the functional incremental type but instead of adjusting the difficulty of the task, the robot is gradually moved to more complex environments.
	
	\item \textbf{Competitive and co-competitive selection}\\
	This kind of selection force robots to compete for the same task (competitive) or for an opposite task (co-competitive, e.g. predator and prey) in the same environment so that even with a static simple fitness function, they are required to evolve more complex skill to beat the opponents.
		
	\item \textbf{Aggregate fitness functions}\\
	These are high-level function that only measure success or failure of the required task so that almost no bias or restrictions are introduced in the evolution. Aggregate fitness functions require the minimum level of \textit{a priori knowledge} but they're also heavily affected by the boostrapping problem, which will be explained in detail in section \ref{sec:bootstrap}.
	
\end{itemize}

\subsection{Searching algorithms}
Once an initial solution has been evaluated, we need to search for new ones in the space of all the possible controllers, that get better fitness function scores.
Usually getting the optimal configuration is out of the question in this field\footnote{Actually, an optmal solution cannot even be defined in this context because the objective function will never define the task entirely: better fitness does't imply better task execution.} due to the large size and low autocorrelation of the search field, hence many euristic search methods are used.
\\
The main block on which many searching algorithms are built is the \textbf{Stochastic Descent} (SD) algorithm that works as follows: first a neighborhood function is defined on the search space, that assign to each solution a set of close alternative configurations. Starting from an initial solution a random neighbor is evaluated, and if it gets a better score than the original configuration, we move to this one and pick another random neighbor to test an so on.
From a search field perspective, what this algorithm do is moving towards a local optima and stopping there.
\\
Many \textbf{trajectory-based} metaeuristics are created from SD that mainly aim to prevent the algorithm from stopping in a local minimum in various ways: accepting worsening steps (\textit{Random Iterative Improvement, Simulated Annealing}), modifying the neighborhood structure (\textit{Variable Neighborhood Search}) or scores (\textit{Dynamic Local Search}), tracking the already explored solutions (\textit{Tabu Search}) or applying a perturbation to the solution and then start SD again (\textit{Iterated Local Search}).
\\
Another way to approach the search is by considering many candidate solutions at once and evolve them together exploiting their interactions; these are called \textbf{population-based} metaeuristics. An example is PSO (\textit{Particle swarm optimization}) where while new local optima are found, particles (that rappresent the current candidate solution swarm) are moved towards them hoping to find a good region of the search landscape. The \textit{Ant Colony Optimization} algorithm makes another good example by using an indirect interaction achieved depositing pheromone in the environment toward which other solutions (\textit{ants}) are attracted.
\\
The most used branch of population-based metaeuristics are \textbf{evolutionary algorithms}, inspired by the Darwinian evolution theory. These are based on three main operators applied to the population: a \textit{selection} process decide which solutions (genotypes) will be used to breed the next generation, then a combination of \textit{crossover} (which recombine two or more different genotypes into new ones) and \textit{mutation} (which is a random alteration of a genotype) operators is used to obtain the new genotypes.

\subsection{Genotype-phenotype mapping}
Many types of evolvable controllers can be expressed in a compact form that only includes their variable parameter values, creating a sort of DNA of a given configuration called \textit{genotype}. In this form, the space of the possible solutions can be better explored and complex operator can be easely applied to a given configuration in a meaningful way (e.g. crossover operator in evolutionary algorithms). The possible encodings for a genotype, as suggested in \citet{art014} are:
\begin{itemize}
	\item Binary encoding
	\item Real-Number encoding
	\item Integer or literal permutation encoding
	\item General data structure encoding
\end{itemize}
A good encoding should also follows four properties that can be summarized as follows:
\begin{itemize}
	\item \textbf{Legality} Any encoding (genotype) permutation corresponds to a solution (phenotype).
	\item \textbf{Completeness} Any solution has a corresponding encoding.
	\item \textbf{Lamarckian Property} the meaning of a subset of the encoding (gene) should be contex-independent.
	\item \textbf{Causality} Small variations on the genotype space due to mutation imply small variations in the phenotype space.
\end{itemize}

\section{Complex Tasks: problem and solutions}
\label{sec:bootstrap}
Complex tasks are one of the main challenges for automatic design of controllers and robot learning. A learning robot accumulate competence on the task from experience (trial and error approach) but when a fitness function points to an objective that is too far beyond its primitive capacities, the evolution will fail to differentiate between initial configurations (that will all be evaluated with the lower score, see \citet{art011}). This problem is know as \textit{bootstrap problem} and its the main cause of failures on complex tasks as explained in \citet{art015}. From a fitness landscape POV this problem is caused by fitness plateus: flat regions that, depending on the search algorithm, can make the evolution very slow or even impossible (\citet{art016}). In the following sections the most used strategies to overcome this problem and to drive the evolution on the right path will be explored.

\subsection{Hierarchial strategy}
This approach is based on a \textit{divide and conquer} way of thiking: to fill the gap between the robot capabilities and the task requirements one can simply teach the robot the needed primitives first, and then evolve it on the complete task exploiting those primitives.
 
\subsubsection{Primitives - Arbitrators}
If the complete task is an immediate combination of a finite number of primitive behaviours, the primitive-arbitrator architecture is the ideal one. Since many of the tasks that appears complex are actually only composite tasks, this approach firstly described in \citet{art006} works very well and its largely adopted in many works.\\

\image{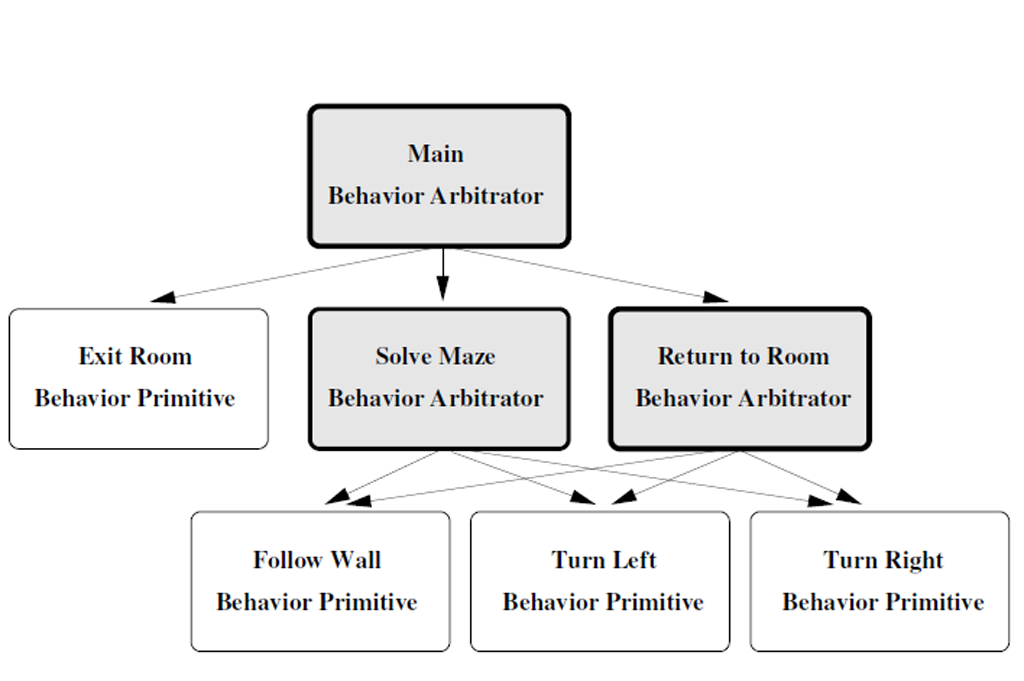}{The controller architecture used in \citet{art002} experiment, composed of 3 behavior arbitrators and 4 behavior primitives.}{0.2}

In \citet{art002} work, the task consisted in a robot that have to: \textit{I.} exit and obstacle filled room to reach a double-T maze, \textit{II.} solve the maze to find another robot using a given light-code indication (so that memory is involved since the robot must remember the code to find the correct path), \textit{III.} resque the other robot by bringing it back to the first room. The choosen architecture that achieved a very high solve-rate (92\%, 22 robots out of 24) can be seen in figure \ref{fig:duarte-arch}, where each block represents a separately evolved ANN. In order to achieve succesful results they manually decomposed the task and used a behavioural fitness function for evolving the main arbitrator.
\\
A similar approach is used in \citet{art004} where arbitrators are explicitly thought as \textit{switchers} that activate one of the sub-modules based on the current sensor inputs. The architecture is proven to be robust to sensor noise by testing it on a real robot assembled with LEGO Mindstorms.

\subsubsection{GP and regulatory genes}
The weakness of an hierarchial strategy lies in the manual task decomposition step, that requires human intervention in the evolutive process. In \citet{art017} a proposed solution is to use GP (particularly \textit{Gene Expression Programming}) with the incorporation of a \textit{regulatory gene} as a part of the chromosome. This gene will regulate the activation of the others, thus composing a layered architecture that can be automatically evolved. This architecture is used in the paper to train robots on wall following and foraging tasks where specialized genes that are activated to perform sub-behaviours can be observed. However the tasks used here were quite simple and the proposed approach is not outperforming similar techniques that don't use regulatory genes.

\subsection{Incremental strategy}
Instead of decomposing the whole task into smaller primitives, here the idea is to learn a simpler and/or partial version of the task, gradually achieving the original goal by acting on the fitness function.

\subsubsection{Training on a simplified task first}
This strategy also know as \textit{scaffolding} or \textit{robot shaping} has been succesfully applied to solve complex tasks. As a first example we can look at \citet{art025} work were a prey-predator task is involved. The predator neural network is firstly evolved against a full capable prey (capable of making many (time-discrete) moves with fast speed) showing that the evolution stall at a low fitness value due to the complexity of the task. Then a set of eight increasingly difficult tasks is set up, where in the first task the prey is completely still (easy to capture) and the last task is the same as the original one. Starting the evolution from the first task, and then gradually moving to the more difficult ones once the robot solves them showed to be effective: every time the task is switched there is a drop in the fitness value and the robot start learning the new task, re-gaining an high fitness even on the last (original) one.
\\
Another more complex experiment is the \citet{art018} one where the scaffolding technique is applied twice for the same task along two dimensions. Here the task is a phototaxis behavior for legged robots that with the direct approach gives good results only after 30 CPU hours. To speed up the task, scaffolding is applied on both an environmental (as different orientations of the light) and morphological (from legless to legged) level in various sequences, even interleaved. The configuration where morphological scaffolding was applied first, showed the best results reaching a good fitness in less than 10 hours.

\subsubsection{Training on a part of the task first}
When we approach a composite task, where there are multiple behaviors to be learnt, an incremental strategy is also possible where these are taught one by one until the robot learn the complete task; this strategy is also known as \textit{behaviour chaining}.
\\
In \citet{art021} an experiment on learning two conflicting tasks is conducted with the goal of understanding what training sequence gives the best results and why. The two analyzed tasks are performed in a virtual arena and are a gradient following behaviour where the robot have to reach the top of the arena, and a rough terrain avoidance where it must learn to avoid ``dangerous'' zones that are arranged in a maze-like configuration. The tests are based on three setups where in the first the robot is trained in both tasks simultaneously, while in the others one of the tasks is learned first, and only at this point the other one is introduced in the fitness function. The results show that using an incremental strategy can lead to better performances if the hardest task is trained first.
\\
In \citet{art022} paper the performance of learning 3 tasks simultaneously is tested against an incremental approach. 
The tasks are namely basic locomotion (since the robot morphology is unknown and the tests are carried out on twelve different robots), turning toward a moving target point and obstacle avoidance. In the incremental strategy the robot is trained in order in each task separately and as the paper shows, every robot configuration reach a good level of fitness faster than in the simultaneous approach.

\subsubsection{Dynamic fitness functions}
Incremental strategies seen so far change environment, robot or fitness function for a discrete number of training steps. The fitness function can be however made time-dependent in a way that the difficulty of the task increase dinamically over the generations without any abrupt change. In \citet{art011} this principle is applied to the evolution of a predator robot in a prey-predator task where the difficulty of the task depends on two parameters: the prey speed and initial distance from the predator. These are modulated by a monotonically time-dependent increasing function $G(t)$ so that the task become more and more difficult until reaching the desired final task. In the paper $G(t)$ is hand-tuned and the results show the increase in stability and the fast convergence against the standard approach.

\subsubsection{Coevolutionary approach}
In the previous example with dynamic fitness functions, the function itself was hand-tuned, thus requiring a human intervention to obtain the best performance. But as shown in \citet{art011} the difficulty level can be also automatically tuned by the algorithm itself exploiting coevolution. This can be achieved by making the reward for solving a task inversely proportional to the number of solutions already found by other robots, forcing the evolution to focus on tasks that are on the edge of what the current population can do. This methodology required more generations of the hand-tuned one to achieve the same fitness but still performed much better than the standard approach.

\subsection{Structural controller evolution}
\label{sec:netevolution}
The complexity of searching a good controller configuration for the task can also be reduced by using a smaller controller thus reducing the search space. But often if the task is a composed/complex one, a small controller is not enought to produce good results. An idea to combine the benefit of both sizes is to start with population of small controllers that due to the reduced number of possible configurations are more likely to give a good start to the evolution that have a lower chance of getting stucked in fitness plateus. Then to get more complex dynamics able to achieve a good performance on the task, another operator is added to the evolutionary process that expand the controller size with a given probability (e.g. adding a state to an FSM or a line of code to an evolved program).
\\
This metodology is mostly applied with ANNs, where a node is added to the network. An example can be found in the NERO game (\citet{art003}) where the rtNEAT algorithm is used: the algorithm is fast enoght to evolve controllers for a swarm of fighters (simulated agents with moving/shooting capabilities) in less than a minute of computation.
 
\subsubsection{Macro elements and mixed evolution}
Operators can also be added to expand the controller with different elements, that are not necessarily omogeneous with the rest of the architecture: these could be used to introduce primitive parametric blocks that already expose a specific non-trivial behaviour. As an example we can look at \citet{art010} paper where the NEAT algorithm is expanded to choose between various operator that can also add radial basis functions (RBF) instead of neurons. This strategy helped the NEAT algorithm doing well in many diferent tasks making it more generic.
 
\subsection{Modular architectures}
Complexity can be also found in the robot itself when its attuators need to be controlled together in a specific way to achieve a primitive behaviour. Think about a task as simple as going straght from point $A$ to point $B$: if the robot is an \textit{e-puck} you just need to activate its two wheels at the same speed to make it go straight. In \citet{art008} a similar task is studied for a miniature helicopter where the sensor inputs are position, velocity, rotation and rotational speed along the three axis for a total of twelve values and the four attuators control the rotor speeds and blade inclinations as in a real helicopter. To stabilize the helicopter and make it able to move forward these attuator needs to be operated in a given, coordinated way, while the other configurations would make it unstable and unable to move along the path giving the worst fitness score: this is exactly what generate the boostrap problem. One of the working approach found in the paper was to split the ANN controlling the robot in four simpler modular networks, each controlling one attuator and using only the needed subset of inputs. Even if these were still evolved together with the same fitness function, this setup was able to yeld much better results.

\subsection{Evolutionary Bias}
As an addition to all the strategies above, evolutionary biases should be mentioned. This is the simplest way to push learning forward in an early stage simply by introducing a tailored term in the objective function that, even if its not directly related to the task itself, can be used drive the evolution on a given path.\\
To show the benefits of a simple bias we should look at \citet{art019} paper where robots using ANNs have to be trained in playing a \textit{capture the flag} game. The complexity of this task lies in sensorial input and network size: a matrix of 150 colors from an image sensor is used, processed with a neural network with up to 5000 connections. An aggregate fitness function evaluates the number of victories, while to overcome the initial bootstrap problem another tailored term is introduced, that measure the distance traveled in the arena by the robot.\\
It's important to remember that biases despite being useful in an initial stage, greatly limit the search space and should thus be used with caution. In the above example the score given by the traveled distance term was at most half of the one given for a victory, so that once a winning behaviour is reached, the bias will be ignored.

\section{Conclusions}
The automatic design of robot controllers is not a fresh topic and many researchers have been exploring this field developing many different evolutionary setups. However to increase the value of these methodologies much work have still to be done in making them capable of tackling increasingly complex tasks while improving their generality.\\
In this paper the most widely used evolutionary setups are explained together with one of the problem that limit their usability on complex tasks: the bootstrap problem. The most used strategies to overcome this problem has been classified showing some of the work done for each one. Despite achieving good results, these techniques either still lack in generality because of their application on a limited number of tasks, or require too many hand-tuned steps (like the task decomposition step in most of the hierarchial approaches), keeping all options open for new works on other strategies or improvements of the existing ones.

\bibliographystyle{plainnat}
\bibliography{matteini2014}

\begin{thebibliography}{22}
\providecommand{\natexlab}[1]{#1}
\providecommand{\url}[1]{\texttt{#1}}
\expandafter\ifx\csname urlstyle\endcsname\relax
  \providecommand{\doi}[1]{doi: #1}\else
  \providecommand{\doi}{doi: \begingroup \urlstyle{rm}\Url}\fi

\bibitem[Amaducci(2011)]{art012}
Matteo Amaducci.
\newblock Design of boolean network robots for dynamics tasks.
\newblock \emph{Seconda Facoltà di Ingegneria, Università di Bologna}, 2011.

\bibitem[Bongard(2011)]{art018}
Josh~C. Bongard.
\newblock Morphological and environmental scaffolding synergize when evolving
  robot controllers.
\newblock \emph{GECCO '11: Proceedings of the 13th annual conference on Genetic
  and evolutionary computation}, 2011.

\bibitem[Busch et~al.(2002)Busch, Ziegler, Aue, Ross, Sawitzki, and
  Banzhaf]{art020}
Jens Busch, Jens Ziegler, Christian Aue, Andree Ross, Daniel Sawitzki, and
  Wolfgang Banzhaf.
\newblock Automatic generation of control programs for walking robots using
  genetic programming.
\newblock \emph{Genetic Programming, Lecture Notes in Computer Science Volume
  2278, 2002, pp 258-267}, 2002.

\bibitem[Christensen and Dorigo(2006)]{art009}
Anders~Lyhne Christensen and Marco Dorigo.
\newblock Evolving an integrated phototaxis and hole-avoidance behavior for a
  swarm-bot.
\newblock \emph{IRIDIA, Universit´e Libre de Bruxelles, Belgium}, 2006.

\bibitem[Doncieux and Mouret(2014)]{art016}
Stephane Doncieux and Jean-Baptiste Mouret.
\newblock Beyond black-box optimization.
\newblock \emph{Evolutionary Intelligence, Volume 7, Issue 2 , pp 71-93}, 2014.

\bibitem[Duarte et~al.(2012)Duarte, Oliveira, and Christensen]{art002}
Miguel Duarte, Sancho Oliveira, and Anders~Lyhne Christensen.
\newblock Hierarchical evolution of robotic controllers for complex tasks.
\newblock \emph{Instituto de Telecomunicaciones \& Instituto Universitario de
  Lisbona (ISCTE-IUL), Lisbon, Portugal}, 2012.

\bibitem[Francesca et~al.(2014)Francesca, Brambilla, Brutschy, Trianni, and
  Birattari]{art001}
Gianpiero Francesca, Manuele Brambilla, Arne Brutschy, Vito Trianni, and Mauro
  Birattari.
\newblock Automode: A novel approach to the automatic design of control
  software for robot swarms.
\newblock \emph{Springer Science+Business Media New York}, 2014.

\bibitem[Garattoni(2011)]{art013}
Lorenzo Garattoni.
\newblock Advanced stochastic local search methods for automatic design of
  boolean network robots.
\newblock \emph{Seconda Facoltà di Ingegneria, Università di Bologna}, 2011.

\bibitem[Gen and Cheng(2000)]{art014}
Mitsuo Gen and Runwei Cheng.
\newblock Genetic algorithms and engineering optimization.
\newblock \emph{John Wiley \& Sons}, 2000.

\bibitem[Gomez and Miikkulainen(1996)]{art025}
Faustino Gomez and Risto Miikkulainen.
\newblock Incremental evolution of complex general behavior.
\newblock \emph{Adaptive Behavior}, 1996.

\bibitem[Haith et~al.(1999)Haith, Colombano, and Jason D.~Lohn]{art011}
Gary~L. Haith, Silvano~P. Colombano, and Dimitris~Stassinopoulos Jason D.~Lohn.
\newblock Coevolution for problem simplification.
\newblock \emph{Proceedings of the Genetic and Evolutionary Computation
  Conference}, 1999.

\bibitem[Kohl and Miikkulainen(2012)]{art010}
Nate Kohl and Risto Miikkulainen.
\newblock An integrated neuroevolutionary approach to reactive control and
  high-level strategy.
\newblock \emph{IEEE TRANSACTIONS ON EVOLUTIONARY COMPUTATION, VOL. 16, NO. 4},
  2012.

\bibitem[Larsen and Hansen(2005)]{art004}
Tobias Larsen and Soren~Tranberg Hansen.
\newblock Evolving composite robot behaviour - a modular architecture.
\newblock \emph{"Roskilde University Computer Science Roskilde, DK-4000"},
  2005.

\bibitem[Lee(1998)]{art006}
Wei-Po Lee.
\newblock Evolving complex robot behaviors.
\newblock \emph{"Research Lab. 301, Microelectronics and Information Systems
  Research Center, National Chiao Tung University, Hsin-Chu, Taiwan, ROC"},
  1998.

\bibitem[Mouret and Doncieux(2009)]{art015}
Jean-Baptiste Mouret and Stephane Doncieux.
\newblock Overcoming the bootstrap problem in evolutionary robotics using
  behavioral diversity.
\newblock \emph{Evolutionary Computation, 2009. CEC '09. IEEE Congress on},
  2009.

\bibitem[Mwaura and Keedwell(2014)]{art017}
Jonathan Mwaura and Ed~Keedwell.
\newblock Evolving robot sub-behaviour modules using gene expression
  programming.
\newblock \emph{Genetic Programming and Evolvable Machines}, 2014.

\bibitem[Nelson(2003)]{art019}
Andrew~L. Nelson.
\newblock Competitive relative performance and fitness selection for
  evolutionary robotics.
\newblock \emph{DEPARTMENT OF ELECTRICAL AND COMPUTER ENGINEERING North
  Carolina State University}, 2003.

\bibitem[Nelson et~al.(2008)Nelson, Barlow, and Doitsidis]{art007}
Andrew~L. Nelson, Gregory~J. Barlow, and Lefteris Doitsidis.
\newblock Fitness functions in evolutionary robotics: A survey and analysis.
\newblock 2008.

\bibitem[Nicolay et~al.(2014)Nicolay, Roli, and Carletti]{art021}
D.~Nicolay, A.~Roli, and T.~Carletti.
\newblock Learning multiple conflicting tasks with artificial evolution.
\newblock \emph{Advances in Artificial Life and Evolutionary Computation},
  2014.

\bibitem[Renzo De~Nardi et~al.(2006)Renzo De~Nardi, Holland, and Lucas]{art008}
Julian~Togelius Renzo De~Nardi, Owen~E. Holland, and Simon~M. Lucas.
\newblock Evolution of neural networks for helicopter control: Why modularity
  matters.
\newblock \emph{"Department of Computer Science University of Essex, UK"},
  2006.

\bibitem[Rossi and Eiben(2014)]{art022}
C.~Rossi and A.~E. Eiben.
\newblock Simultaneous versus incremental learning of multiple skills by
  modular robots.
\newblock \emph{Evolutionary Intelligence August 2014, Volume 7, Issue 2, pp
  119-131}, 2014.

\bibitem[Stanley et~al.(2005)Stanley, Bryant, and Miikkulainen]{art003}
Kenneth~O. Stanley, Bobby~D. Bryant, and Risto Miikkulainen.
\newblock Real-time neuroevolution in the nero video game.
\newblock \emph{Department of Computer Sciences University of Texas at Austin,
  TX 78712 USA}, 2005.

\end{thebibliography}

\end{document}